%% file: root.tex
\documentclass[letterpaper, 10 pt, conference]{ieeeconf}  

\IEEEoverridecommandlockouts                              

\usepackage{graphicx}
\usepackage{amsmath} 
\usepackage{amssymb}  
\usepackage{xcolor}
\usepackage{url}
\usepackage{siunitx}
\usepackage{booktabs}
\usepackage{subcaption}

\title{\LARGE \bf
Probabilistic Physics Digital Twin for Human-Robot Interaction
}

\author{%
Yushen Hu$^{1,*}$,
Harshvardhan Chandirasekar$^{1,*}$,
Apurv Kushwaha$^{1}$,
Jonathan King$^{2}$,
and Zhu-Tian Chen$^{1}$\\[0.5em]
{\small $^{1}$University of Minnesota, USA}\\
{\small $^{2}$Howard University, USA}\\
{\small Correspondence: \texttt{hu000809@umn.edu}}
\thanks{$^{*}$Equal contribution.}%
}

\begin{document}

\maketitle
\thispagestyle{empty}
\pagestyle{empty}

\begin{abstract}

Digital twins are increasingly used in robotics to simulate interactions, predict action outcomes, and support planning. However, constructing accurate digital twins requires precise estimates of physical properties such as mass, friction, and center of mass, which are often difficult to obtain in real-world environments. We propose probabilistic digital twins, which represent uncertain physical properties as probability distributions rather than fixed values. By simulating actions across these distributions, the twin predicts a range of possible outcomes and their likelihoods. These probabilistic simulations can guide both human and robot actions by identifying actions that are effective despite physical uncertainty. We demonstrate this framework through human–robot interaction scenarios, including interfaces that communicate simulated outcomes to users and robot planning strategies that explicitly account for uncertainty.

\end{abstract}

\input{tex/1-intro}

\input{tex/2-relatedworks}

\input{tex/3-method}

\input{tex/4-implementation}

\input{tex/5-evaluation}

\input{tex/6-discussion-limitation}

\bibliographystyle{IEEEtran}
\bibliography{references}

\end{document}

%% file: tex/1-intro.tex
\section{Introduction}

Digital twins provide digital representations of real-world environments that are kept aligned with observations of the physical scene \cite{grieves2016digital}. In robotics, a digital twin can represent not only what is currently in the environment, but also what may happen under candidate actions. When coupled with a rigid-body physics engine, for example, the twin can simulate how a scene evolves under gravity, contact, collision, and friction, enabling a robot to evaluate possible physical consequences before acting in the real world.

However, such prediction requires more than accurate scene geometry. A physics simulator also requires properties such as mass, friction, restitution, and center of mass. Unlike object geometry and pose, which can increasingly be estimated from RGB or RGB-D observations, many of these properties are only partially observable from a static scene \cite{xu2019densephysnet}. A conventional digital twin must nevertheless assign them specific values and therefore predicts action outcomes under one particular estimate of the scene's physical properties.

This can be problematic near physical decision boundaries. A placement that is stable under one plausible friction coefficient may slide under another, while a stack that remains upright under one plausible center of mass may topple under a slightly different one. In such cases, improving a point estimate does not necessarily resolve the ambiguity because the available observations may not uniquely determine the underlying physical properties. We therefore ask whether a digital twin can preserve this uncertainty and propagate it through physics simulation rather than committing prematurely to a single physical model.

We introduce a \textbf{Probabilistic Physics Digital Twin for Human-Robot Interaction}, in which uncertain physical properties are represented as distributions over plausible values. Instead of simulating each candidate action under a single estimated physical model, the twin samples multiple plausible physical realizations of the observed scene and evaluates the action across the resulting simulated futures. The collection of rollouts therefore captures a distribution of physically plausible consequences rather than a single predicted outcome.

Importantly, the probabilistic twin does not need to determine which physical realization is exactly correct. Instead, each simulated future can be summarized using a task-specific \emph{physical consequence}, and that consequence is aggregated across plausible realizations. For object placement, the consequence may measure instability and impact severity. For actions that induce motion, such as pushing an object, the consequence may instead describe the distribution of locations that the object may reach. This converts uncertainty over latent physical properties into an action-level estimate of expected physical consequence, allowing downstream decisions to explicitly account for physical uncertainty.

In this work, we formulate a general probabilistic physics pipeline for Digital Twins, in which uncertainty over physical properties is propagated through forward simulation to estimate task-relevant physical consequences. The framework supports different forms of physical reasoning and provides spatial representations that can be used by downstream human and robotic systems. We instantiate the full pipeline using object-placement stability, where the resulting consequence representation supports both human guidance and robotic action selection. Our empirical evaluation therefore focuses on the placement-stability instantiation. Our contributions are as follows:

\begin{itemize}

    \item We introduce a probabilistic digital-twin formulation that represents uncertain physical properties as distributions and propagates this uncertainty through forward physics simulation.

    \item We formulate task-specific probabilistic consequence representations within this framework and construct spatial representations that support multiple forms of physical reasoning, including downstream human guidance and robotic action selection.

    \item We instantiate and evaluate the framework for object-placement stability through a user study and a robotic ablation study.

\end{itemize}

%% file: tex/2-relatedworks.tex
\section{Related Works}

\subsection{Probabilistic Programming}

Probabilistic programming represents structured uncertainty within executable
generative models. Church formalized stochastic generative processes as
programs supporting general-purpose probabilistic inference
\cite{goodman2012church}, establishing probabilistic programs as compositional
world models for prediction and decision making under uncertainty.

This perspective has been influential in computational models of intuitive
physics. Battaglia et al.'s intuitive physics engine samples uncertain scene
properties and forces, simulates the resulting physical worlds, and aggregates
outcomes to predict human physical judgments \cite{battaglia2013simulation};
Ullman et al.\ develop this into a game-engine-like architecture for physical
reasoning \cite{ullman2017mind}. Both treat simulation not as a deterministic
forward model, but as a mechanism for reasoning over possible physical worlds.

Recent work links such executable world models to natural language. PiLoT maps
linguistic descriptions to probabilistic programs, samples consistent scenes,
and evaluates them with a physics simulator \cite{zhang2023grounded}; Wong et
al.\ generalize this idea, using language models to translate natural language
into a probabilistic language of thought whose programs interface with symbolic
modules including physics simulators and planners \cite{wong2023word}.

\subsection{Digital Twins}

A useful Digital Twin requires reconstructing both the observable state of the
physical environment and the latent properties that govern its behavior. Recent
work has substantially improved scene and object reconstruction using
representations such as Neural Radiance Fields (NeRFs) and 3D Gaussian Splatting
(3DGS), including systems designed specifically for robotic manipulation and
planning \cite{sun2026high,shorinwa2024splat,li2026realtwin}. Other approaches
augment reconstructed geometry with physical behavior or directly estimate
simulation-relevant properties. PhysGaussian couples physical dynamics with 3D
Gaussian representations \cite{xie2024physgaussian}, while PAC-NeRF,
PhysDreamer, and Physics3D estimate physical or material properties from visual
observations and dynamics \cite{li2023pac,zhang2024physdreamer,liu2024physics3d}.
More recent systems such as PhysX-Anything and D-REX target simulation-ready
physical representations and system identification for downstream interaction
\cite{cao2026physx,lou2026d}. Together, these methods improve the fidelity of
the reconstructed physical model, but typically produce a particular estimated
realization of the scene.

A complementary line of work explicitly represents uncertainty within Digital
Twins. Probabilistic Digital Twins model uncertain system states or parameters
as distributions and propagate this uncertainty for prediction, inference, and
decision making \cite{kapteyn2021probabilistic, kochunas2021digital,
agrell2023optimal}. Such formulations have been explored in domains including
aerospace, structural monitoring, and engineering decision support, where
uncertainty is updated or queried as new observations become available.

Our work follows this probabilistic view but focuses on interactive rigid-body
environments. Rather than requiring uncertain physical properties to be resolved
before the Digital Twin can be used, we retain uncertainty over properties such
as mass, friction, restitution, and center of mass and propagate it through
forward physics simulation to obtain task-relevant consequence representations
for human and robotic decision making.

%% file: tex/3-method.tex
\section{Method}
\label{sec:method}

In this section, we describe how probabilistic physics is formulated to answer task-relevant queries in human--robot interaction. Such queries may ask, for example, ``How safe is it to place an object at this location?'' or ``Where is an object likely to move after being pushed?'' We distinguish between two forms of consequence: scalar consequences, which assign a value to a candidate query, and spatial consequences, which directly produce a distribution of possible
locations.

\subsection{Probabilistic Formulation}

Let $s$ denote the observed scene, $a\in\mathcal{A}$ a candidate action, and $\boldsymbol{\theta}\in\Theta$ the uncertain physical parameters. We represent the physical properties consistent with the observation by $p(\boldsymbol{\theta}\mid s)$. For one realization $\boldsymbol{\theta}$, the physics simulator produces a trajectory

\begin{equation}
\tau =
\mathrm{Sim}(s,a;\boldsymbol{\theta}).
\label{eq:simulation}
\end{equation}

Rather than committing to one physical realization, the probabilistic twin evaluates the action across samples from $p(\boldsymbol{\theta}\mid s)$. The resulting trajectories can then be interpreted either as scalar consequences or as spatial consequences, depending on the downstream query. A deterministic digital twin is the special case in which $p(\boldsymbol{\theta}\mid s)$ collapses to a point estimate. Additional uncertain inputs, such as external disturbances, can be sampled in the same manner.

\subsection{Scalar Consequence Queries}

For some tasks, each simulated trajectory can be summarized by a scalar task-specific consequence. Examples include placement stability and impact severity at a candidate query.

\subsubsection{Consequence functional}

\begin{figure}[t]
    \centering

    \begin{subfigure}[b]{0.48\linewidth}
        \centering
        \includegraphics[width=\linewidth]{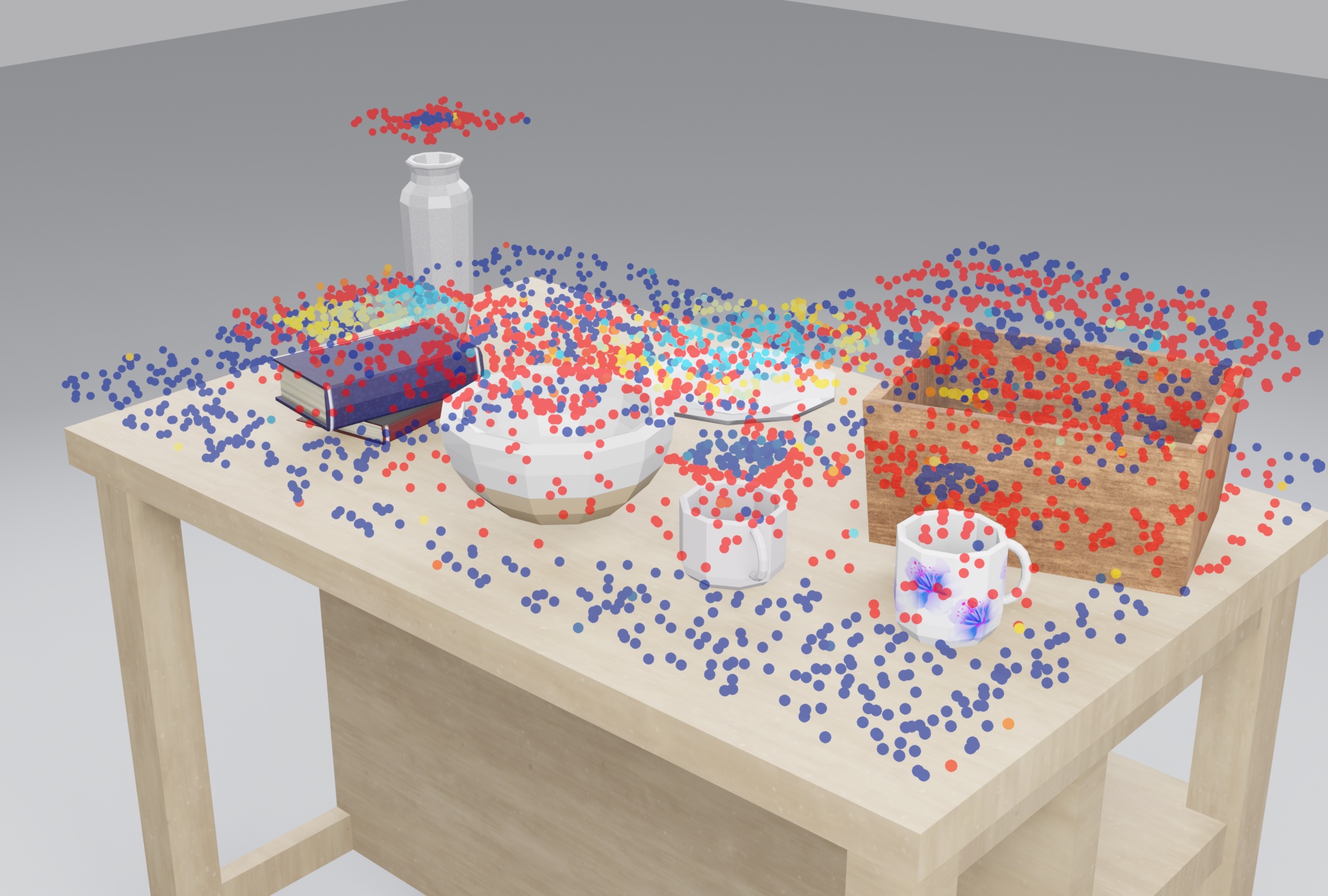}
        \caption{Stability sampling}
        \label{fig:stability_sampling}
    \end{subfigure}
    \hfill
    \begin{subfigure}[b]{0.48\linewidth}
        \centering
        \includegraphics[width=\linewidth]{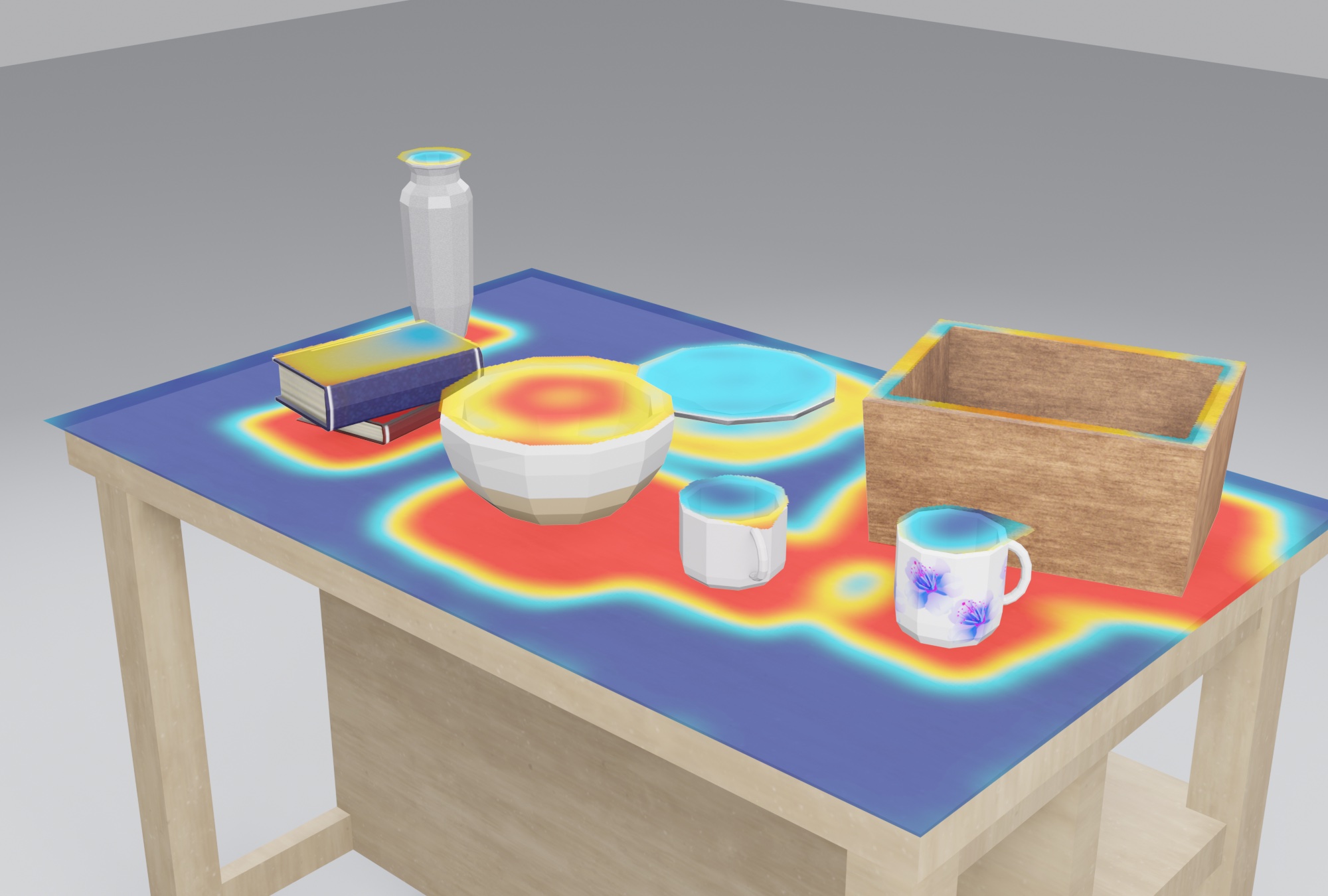}
        \caption{Stability field}
        \label{fig:stability_field}
    \end{subfigure}

    \vspace{0.4em}

    \begin{subfigure}[b]{0.48\linewidth}
        \centering
        \includegraphics[width=\linewidth]{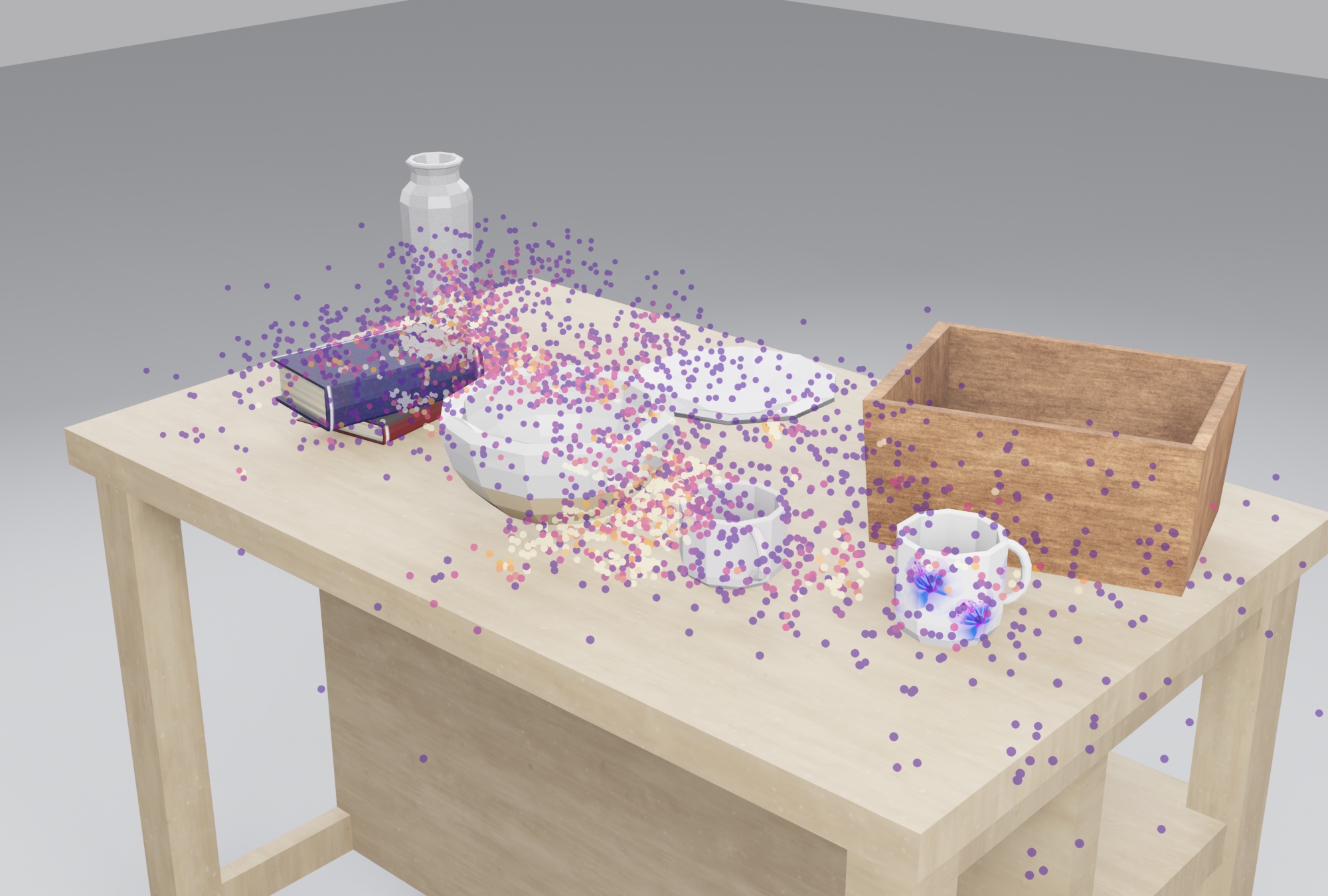}
        \caption{Impact sampling}
        \label{fig:impact_sampling}
    \end{subfigure}
    \hfill
    \begin{subfigure}[b]{0.48\linewidth}
        \centering
        \includegraphics[width=\linewidth]{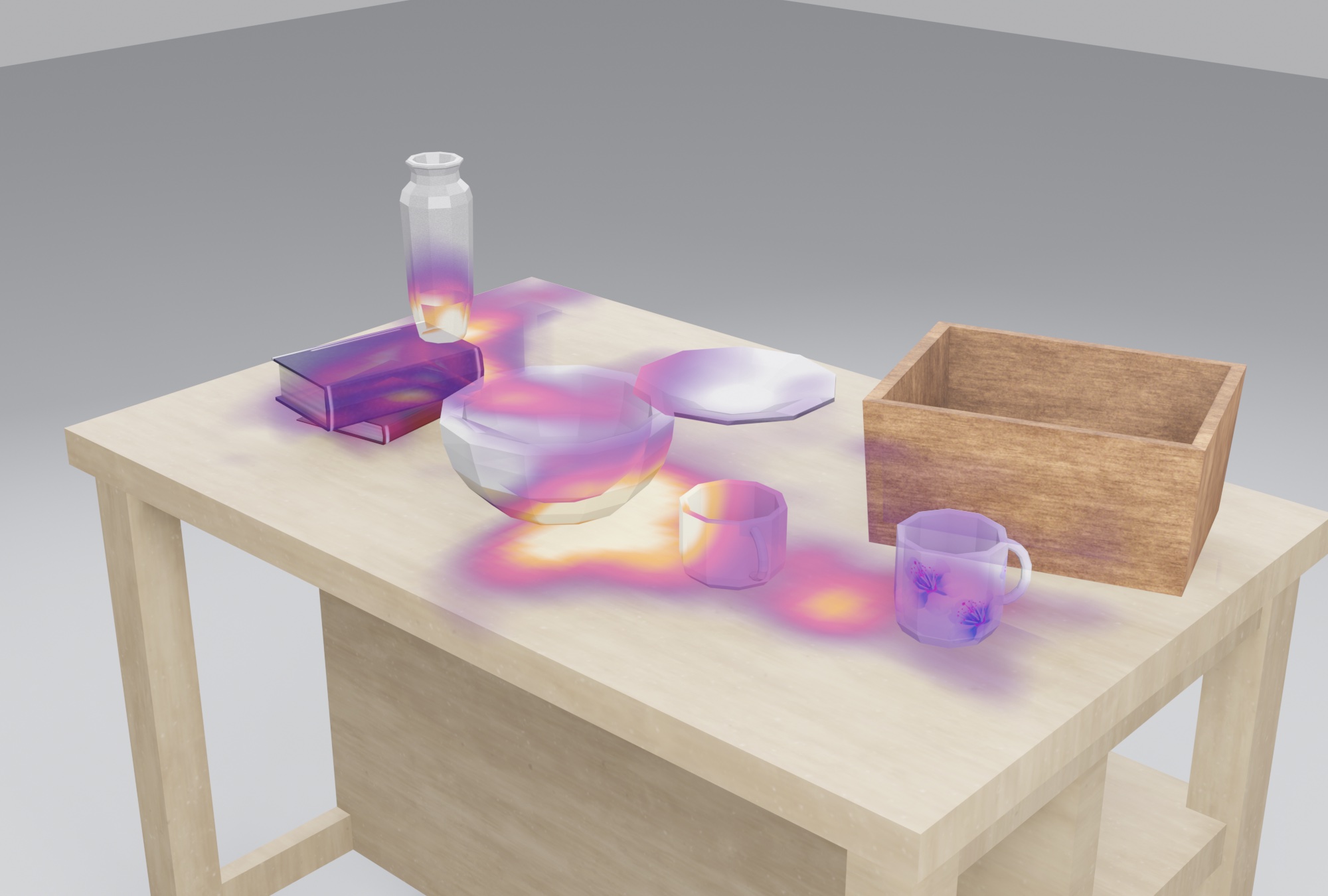}
        \caption{Impact-risk field}
        \label{fig:impact_field}
    \end{subfigure}

    \vspace{0.4em}

    \caption{
    Visualization of scalar consequence-field construction. (a) and (b) show the stability field, (c) and (d) show the impact field. Each sampled point represents an expected consequence obtained by averaging multiple simulated episodes over sampled physical realizations.
    }
    \label{fig:field_construction}
\end{figure}

Let $\phi_{\mathrm{sc}}$ denote a scalar consequence functional. For a simulated trajectory,

\begin{equation}
\Phi_{\mathrm{sc}}(s,a,\boldsymbol{\theta})
=
\phi_{\mathrm{sc}}
\left(
\mathrm{Sim}(s,a;\boldsymbol{\theta})
\right)
\in\mathbb{R}_{\geq 0}.
\label{eq:scalar_consequence}
\end{equation}

The expected scalar consequence is

\begin{equation}
U(a\mid s)
=
\mathbb{E}_{\boldsymbol{\theta}\sim p(\boldsymbol{\theta}\mid s)}
\left[
\Phi_{\mathrm{sc}}(s,a,\boldsymbol{\theta})
\right].
\label{eq:expectation}
\end{equation}

\paragraph{Placement stability}
For placement stability, the action itself specifies a candidate location $a=\mathbf{x}$. The consequence measures the additional transient motion caused by releasing the object at that location. Let

\begin{equation}
\mathcal{T}(t)
=
\sum_{b=1}^{B}K_b(t)
\label{eq:kinetic}
\end{equation}

denote the total kinetic energy of the $B$ simulated rigid bodies. We compare each candidate placement against a matched baseline rollout. Over a simulation horizon $T$,

\begin{equation}
\Phi_0(\boldsymbol{\theta})
=
\frac{1}{T}
\int_0^T
\mathcal{T}_0(t;\boldsymbol{\theta})\,dt,
\label{eq:baseline}
\end{equation}

and define

\begin{equation}
\Phi_{\mathrm{place}}
(\boldsymbol{\theta},\mathbf{x})
=
\left[
\frac{1}{T}
\int_0^T
\mathcal{T}(t;\boldsymbol{\theta},\mathbf{x})\,dt
-
\Phi_0(\boldsymbol{\theta})
\right]_+ .
\label{eq:excess}
\end{equation}

A lower value indicates that the candidate introduces less additional motion.

\paragraph{Impact risk}

Impact risk provides another scalar consequence. Given a spatial query, a rollout may be summarized by the frequency and severity of nearby collisions. Severity can be characterized using quantities such as contact impulse. The same formulation therefore produces an expected impact-risk value for each candidate query.

For spatial decision tasks, estimating the consequence at a single location is often insufficient. A downstream system may also need to answer follow-up queries such as ``If this location is unsafe, where should the object be placed instead?'' To support such spatial queries, we construct a continuous consequence field that can be queried directly or differentiated to guide local search toward lower-consequence locations.

\subsubsection{Monte Carlo estimation}

Given independent
samples
$\{\boldsymbol{\theta}^{(m)}\}_{m=1}^{M}
\sim p(\boldsymbol{\theta}\mid s)$,

\begin{equation}
\hat{U}(a\mid s)
=
\frac{1}{M}
\sum_{m=1}^{M}
\Phi_{\mathrm{sc}}
(s,a,\boldsymbol{\theta}^{(m)}).
\label{eq:mc}
\end{equation}

For a spatial query task, we evaluate this estimator at a finite collection of
locations $\{\mathbf{x}_i\}_{i=1}^{N}$. Each resulting value
$\hat{U}_i$ therefore represents an expected consequence obtained from multiple
probabilistic rollouts rather than one deterministic simulation.

\subsubsection{Scalar consequence field}
\label{sec:fieldconstruction}

The sampled estimates
$\{(\mathbf{x}_i,\hat{U}_i)\}_{i=1}^{N}$ provide consequences only at the
evaluated locations. To support arbitrary spatial queries and search for nearby
locations with lower expected consequence, we reconstruct them into a
continuous lookup field.

For an arbitrary query $\mathbf{q}$,

\begin{equation}
\tilde{U}(\mathbf{q})
=
\frac{
\sum_i k_h(\mathbf{q}-\mathbf{x}_i)\hat{U}_i
}{
\sum_i k_h(\mathbf{q}-\mathbf{x}_i)
},
\label{eq:spatial_field}
\end{equation}

where $k_h$ is a Gaussian kernel with bandwidth $h$.

We additionally define

\begin{equation}
S(\mathbf{q})
=
\sum_i k_h(\mathbf{q}-\mathbf{x}_i)
\label{eq:field_support}
\end{equation}

as the local sample support and treat queries with
$S(\mathbf{q})<S_{\min}$ as insufficiently supported. Figure \ref{fig:field_construction} illustrates the sampling and construction of the field.

\subsection{Spatial Consequence Queries}

\begin{figure}[t]
    \centering

    \begin{subfigure}[b]{0.48\linewidth}
        \centering
        \includegraphics[width=\linewidth]{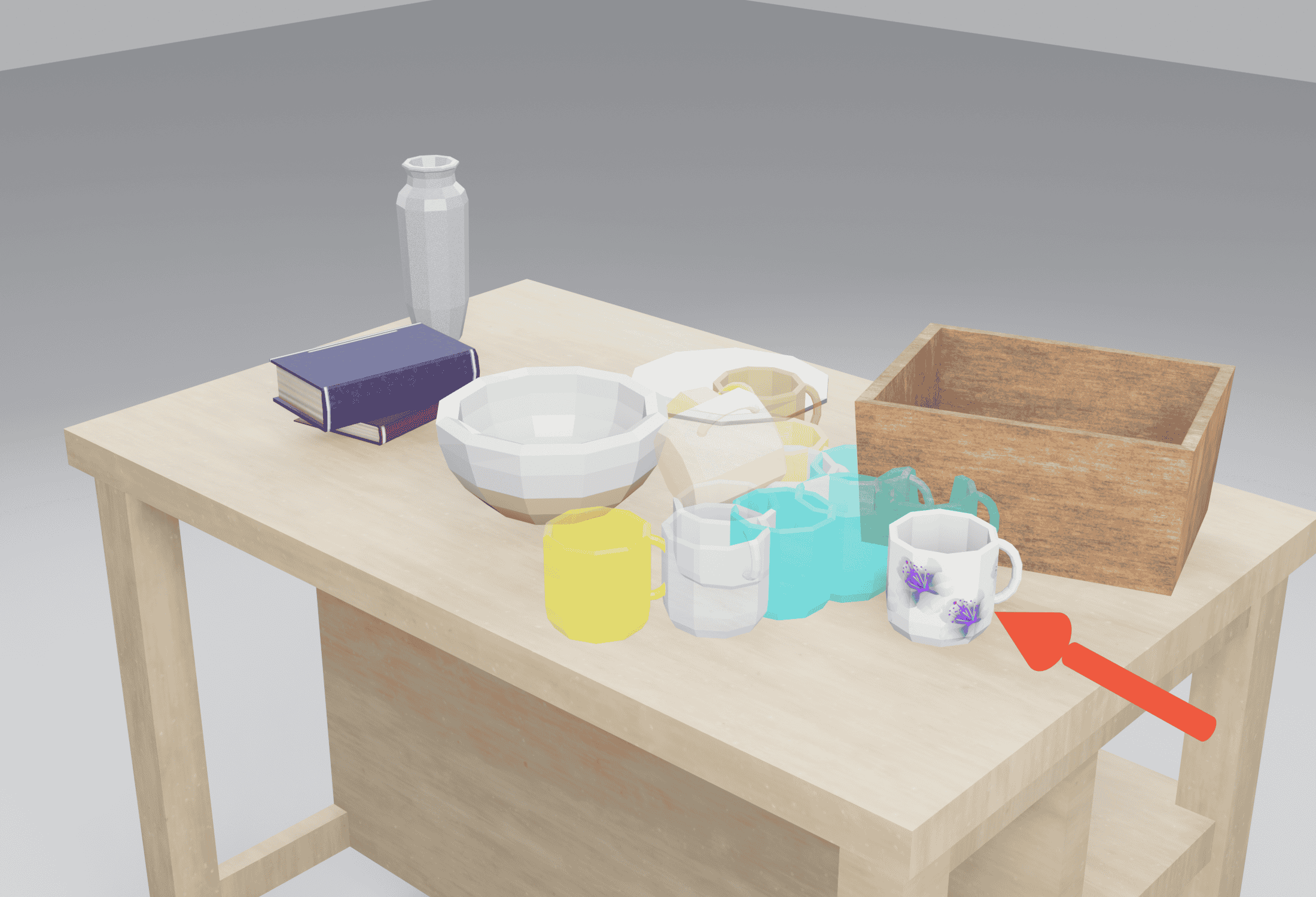}
        \caption{Consequences sampling}
        \label{fig:mug_pose}
    \end{subfigure}
    \hfill
    \begin{subfigure}[b]{0.48\linewidth}
        \centering
        \includegraphics[width=\linewidth]{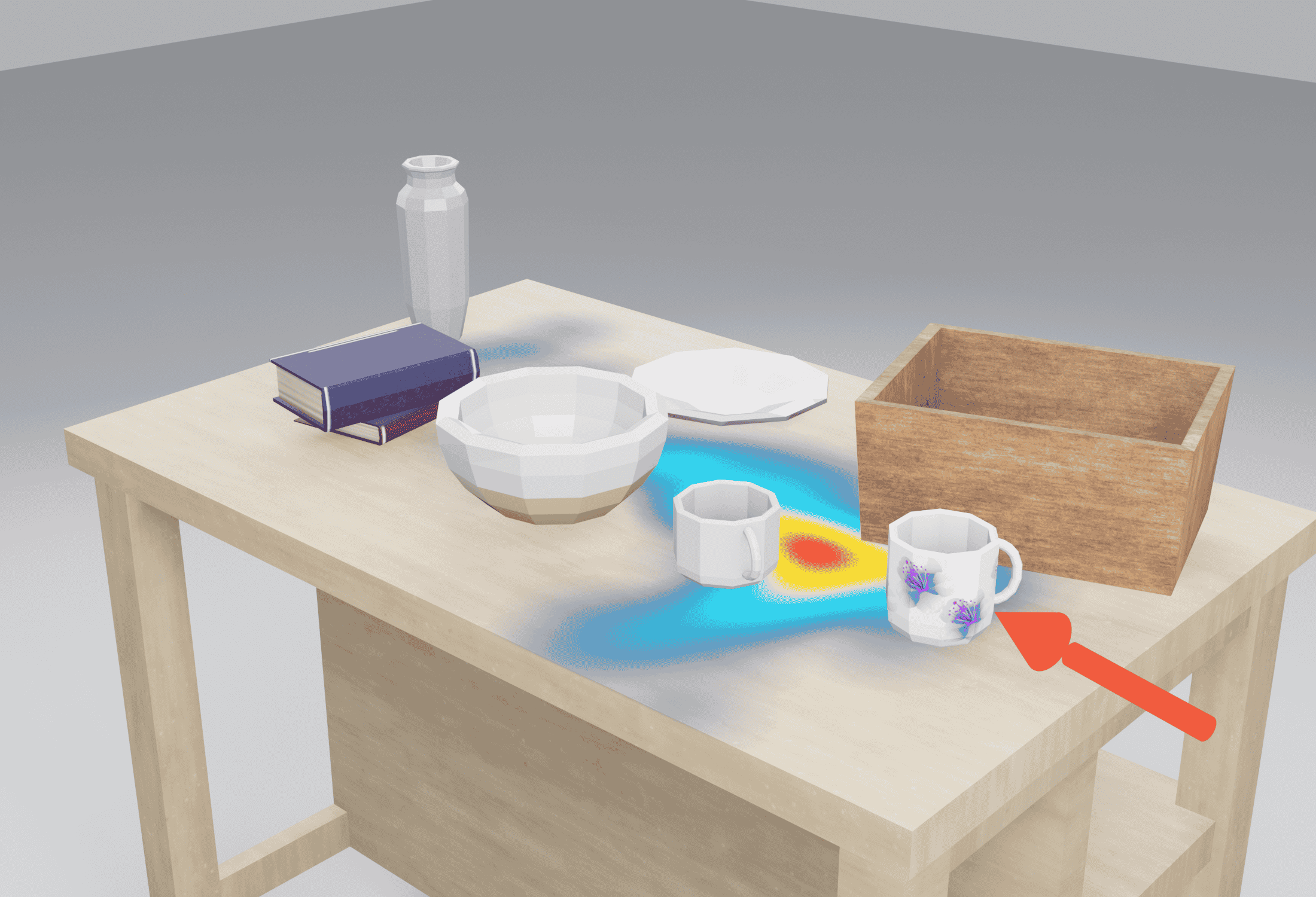}
        \caption{Field construction}
        \label{fig:mug_heat}
    \end{subfigure}

    \vspace{0.4em}

    \caption{
    Visualization of location-valued consequence-field construction. The distribution of locations are sampled and constructed into a field.}
    \label{fig:spatial}
\end{figure}

Other actions produce consequences that are inherently spatial. For example,
after applying a force or impulse to an object, the desired query may be
``Where will the object move to?'' In this case, the simulator does not first
produce a scalar consequence at a chosen query location. Instead, each rollout
directly produces a collection of spatial locations.

\subsubsection{Location-valued consequence functional}

Let a trajectory-derived consequence be represented by the set

\begin{equation}
\mathcal{X}(\tau)
=
\left\{
\mathbf{x}_j
\right\}_{j=1}^{J},
\qquad
\mathbf{x}_j\in\mathbb{R}^{3},
\label{eq:location_consequence}
\end{equation}

where $\mathcal{X}(\tau)$ contains locations occupied or reached by the object
during the simulated trajectory.

For probabilistic reachability, each physical realization therefore generates

\begin{equation}
\mathcal{X}^{(m)}
=
\mathcal{X}
\left(
\mathrm{Sim}(s,a;\boldsymbol{\theta}^{(m)})
\right).
\label{eq:reachability_samples}
\end{equation}

The collection
$\{\mathcal{X}^{(m)}\}_{m=1}^{M}$ represents the distribution of possible
object locations under uncertain physics.

\subsubsection{Spatial field construction}

Because the consequence samples are already locations in $\mathbb{R}^{3}$, a
spatial field can be constructed directly from their distribution. For example,
we define the probabilistic reachability field

\begin{equation}
\hat{R}(\mathbf{q}\mid s,a)
=
\frac{1}{M}
\sum_{m=1}^{M}
\max_{\mathbf{x}_j\in\mathcal{X}^{(m)}}
k_h(\mathbf{q}-\mathbf{x}_j).
\label{eq:reachability_field}
\end{equation}

A high value of $\hat{R}(\mathbf{q}\mid s,a)$ indicates that the object reaches
or passes near $\mathbf{q}$ in many plausible physical realizations.

Unlike the scalar case, the distribution of simulated locations itself
provides the samples from which the spatial field is formed. Figure \ref{fig:spatial} demonstrates the construction of the field.

\subsection{Posterior Refinement}
\label{sec:posterior}

The probabilistic formulation also admits closed-loop refinement through
interaction. Although we do not implement this update in the present system,
the current belief $p(\boldsymbol{\theta}\mid s)$ can in principle be refined
after observing the outcome of an executed action. After executing action $a$
and observing $y^{\mathrm{obs}}$,

\begin{equation}
p(\boldsymbol{\theta}\mid s,a,y^{\mathrm{obs}})
\propto
p(y^{\mathrm{obs}}\mid s,a,\boldsymbol{\theta})
\,p(\boldsymbol{\theta}\mid s).
\label{eq:posterior}
\end{equation}

The resulting posterior can replace the current belief in subsequent
predictions. When a tractable likelihood is unavailable, the update can instead
be approximated through likelihood-free inference by comparing the observed
outcome with outcomes simulated under sampled physical parameters. We leave
implementation and evaluation of this refinement process to future work.

%% file: tex/4-implementation.tex
%

\section{System Instantiation}
\label{sec:implementation}

We instantiate the probabilistic digital-twin framework of
Sec.~\ref{sec:method} for object placement, using placement stability as the
physical consequence of interest. The system maintains a live digital
representation of the observed scene and uses this state as the starting point
for probabilistic forward simulation. For each candidate placement, multiple
physical realizations are generated by sampling uncertain parameters from
$p(\boldsymbol{\theta}\mid s)$, and the placement is simulated independently
under each realization. Their consequences are aggregated according to
Eq.~\eqref{eq:mc} and accumulated into the spatial consequence field of
Sec.~\ref{sec:fieldconstruction}.

This architecture separates two roles of the digital twin. The \emph{live twin}
tracks the currently observed physical scene, while the \emph{simulation
branches} represent plausible futures under uncertain physical properties and
hypothetical actions. The resulting stability field provides a common
representation that can be visualized for a human user or queried by a robot
for placement selection and refinement.

\subsection{Physical Parameter Estimation}

To instantiate the parameter distributions used for probabilistic sampling, we use Qwen 3 \cite{yang2025qwen3} as the primary vision-language model (VLM) for physical-property estimation. Given a snapshot of the scene, the VLM identifies the individual objects and estimates their relevant physical properties based on visual appearance and contextual cues. For each property, the estimated value is used as the mean of the corresponding parameter distribution, while the distribution variance is determined from the model's confidence in that estimate. These distributions provide the initial $p(\boldsymbol{\theta}\mid s)$ used to generate plausible physical realizations during probabilistic simulation.

For the controlled evaluation in Sec. \ref{sec:evaluation}, we instead use a fixed set of objects with predefined physical parameters and uncertainty ranges. This isolates the effect of probabilistic simulation from variability in VLM-based parameter estimation and ensures that all compared conditions operate on the same underlying physical assumptions.

\subsection{Stability field}
\label{sec:impl_stability}

The stability field is implemented with a GPU rigid-body solver (Warp XPBD) that simulates many candidate placements and physical-parameter draws in parallel. Candidate locations are sampled from support surfaces.

Sampled consequences are accumulated into a fixed voxel grid for efficient field reconstruction. Grid resolution determines the spatial resolution of the stored field, while sample count determines how well each region is estimated.

All simulations run on a single NVIDIA GB10 Grace--Blackwell module using Newton 1.5 and Warp 1.16 with an XPBD timestep of $\Delta t=1/240$\,s and ten solver iterations per step. On a nine-body scene with 103 convex hulls, a global batch of $256\times4$ candidate--parameter combinations ($1{,}024$ worlds, $100$ steps) completes in $2.23\pm0.01$\,s, while a focused $64\times4$ batch over a one-second horizon completes in $1.14\pm0.03$\,s. A discrete release check over seven candidate poses and three parameter draws completes in $0.55\pm0.03$\,s.

Because these rates are below frame-coupled interaction speed, field estimation runs asynchronously and streams partial results, refreshing at approximately $0.45$\,Hz globally and $0.88$\,Hz in the focused region. A full $30{,}000$-sample field converges in approximately $4.4$ minutes, while only the $0.55$\,s release check lies on the critical path of discrete action confirmation.

\subsection{User interaction}
\label{sec:impl_interaction}
\label{sec:interface}

\begin{figure}
    \centering
    \includegraphics[width=\linewidth]{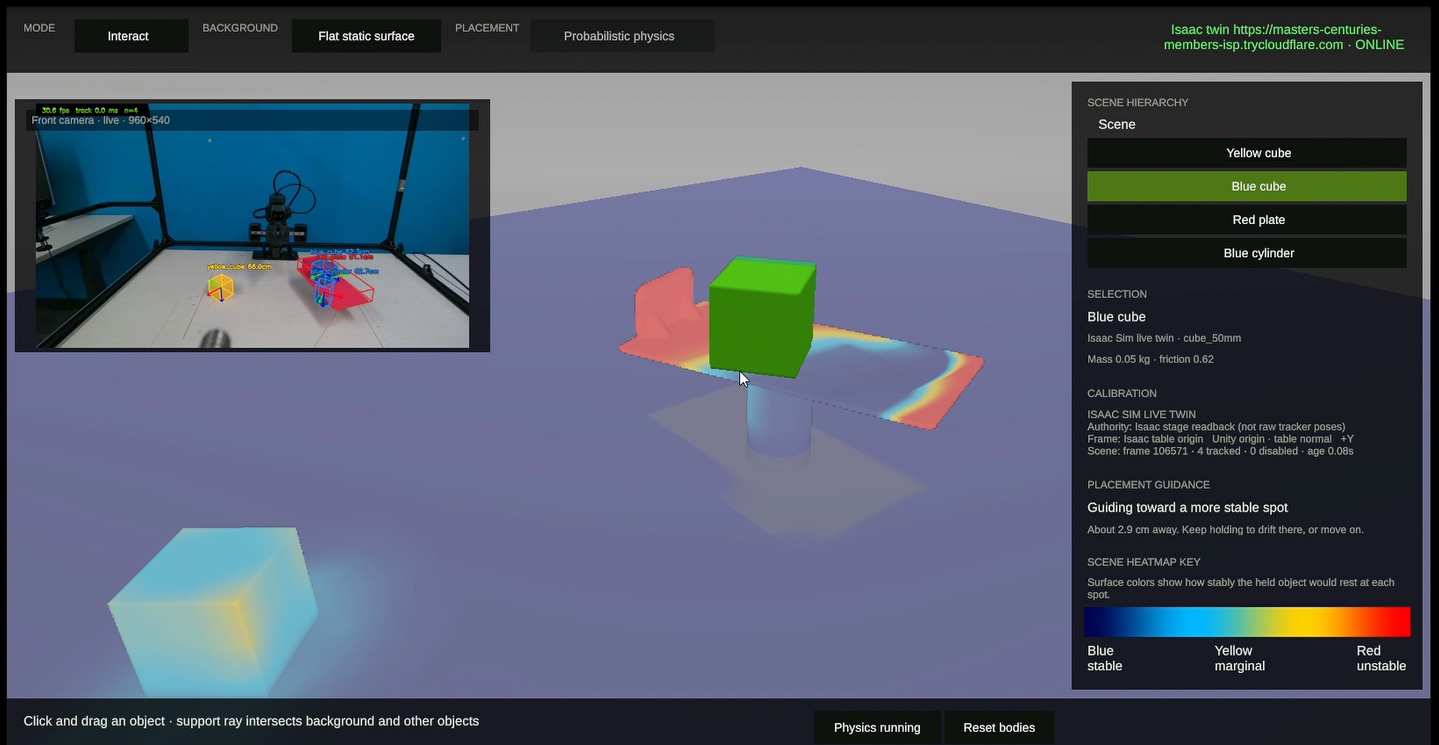}
    \caption{User interface for probabilistic placement guidance. The live camera view
    (upper left) shows the physical scene and tracked objects, while the main view
    displays the corresponding Digital Twin with the reconstructed stability field
    over candidate support surfaces. The field is visualized from stable (blue) to
    unstable (red) and is used to provide placement guidance during interaction.}
    \label{fig:ui}
\end{figure}

The user-facing client is a Unity WebGL application that receives scene geometry and live object poses, streams stability samples from the sampling service, and accumulates them into the voxel grid of Sec.~\ref{sec:impl_stability}. The same grid is used both to visualize the stability field over the scene and to query guidance values during interaction.

Users manipulate an object by pointing to candidate support surfaces in the scene. The selected placement is continuously evaluated against the local stability field, while the current interaction region is communicated back to the sampler to concentrate additional samples where the user is working.

Guidance is applied adaptively according to the user's motion. During deliberate movement, the user retains control over the placement; as motion slows, the system increasingly biases the object toward a nearby lower-consequence location, with the guidance strength reduced in regions of weak sample support. On release, the object is committed to the locally optimized placement suggested by the field.

Figure \ref{fig:ui} displays the user interfaced. The physical scene and its
tracked objects are displayed alongside the corresponding Digital Twin, where
the probabilistic stability field is overlaid on candidate support surfaces.
This visualization exposes the predicted consequence of different placement
locations to the user and provides guidance toward regions of lower expected
instability.

\subsection{Robot workcell}
\label{sec:impl_workcell}

The cell is a 1.80\,m~$\times$~1.60\,m table with a Trossen WidowX~AI six-DoF
arm~\cite{trossen_widowxai} at the centre of its rear edge, observed at 30\,fps
by two RealSense D455 RGB-D cameras~\cite{keselman2017realsense}, one front-on
and one oblique.
Perception, the twin and the arm driver run as separate processes on one
workstation (2$\times$RTX~5090) and are addressed over HTTP.

\paragraph{Perception}
A registry declares each object's metric CAD mesh, symmetry class and HSV
signature, and generates both the detector prompt and the twin scene.
Each pass runs Grounding~DINO~\cite{liu2024groundingdino} and
SAM~2.1~\cite{ravi2024sam2,medeiros_langsam}, assigns masks to registry objects
by linear assignment~\cite{crouse2016assignment} over colour, geometry and
detector evidence rather than trusting detector labels, and registers each
object against its mesh with
FoundationPose~\cite{wen2024foundationpose}.

\paragraph{Planning and execution}
The planner receives one object and the goal pose decided upstream and returns a collision-checked joint trajectory
that moves that object alone, so the robot executes exactly the action that was
committed.
Transfer legs are planned by cuRobo~\cite{sundaralingam2023curobo} around the
table, cage and every tracked object; contact legs are straight-line TCP
motions solved with damped-least-squares IK~\cite{nakamura1986dls} and
collision-checked at every sample, with the grasped object attached to the tool
model.

\textbf{Calibration.} Placement accuracy, not model fidelity, sets the spatial
scale the field must resolve, so we anchor the extrinsics to the arm's own
kinematics: with the arm in gravity compensation, a Kabsch--Umeyama
fit~\cite{kabsch1976,umeyama1991} of forward-kinematic readings at 28 ChArUco
corners gives $T^{\mathrm{rig}}_{\mathrm{board}}$ at
2.5\,mm RMS, and each camera solves the same board
with~IPPE~\cite{collins2014ippe}.
This residual fixes the spatial scale of the simulation study, whose injected
placement error is 4\,mm.

\subsection{Digital twin}
\label{sec:impl_twin}

An Isaac~Sim~5.1 stage~\cite{isaacsim} mirrors the planner's cell.
Every 100\,ms it seats each vision pose on its support and pins the body
kinematically, so the twin tracks observation rather than simulator drift; a
grasped object instead follows the arm's streamed joint states (5\,Hz) until
release.
Poses refresh once per detection pass (2--3\,s), and this perception latency
dominates the real-to-digital delay.
With the pinning off and the arm driven through the real controller under
PhysX, the same stage becomes the robot-in-the-loop simulator of
Sec.~\ref{sec:sim}.

%% file: tex/5-evaluation.tex

\providecommand{\armReq}{\textsc{None}}
\providecommand{\armGeo}{\textsc{Geo}}
\providecommand{\armGeoK}{\textsc{Geo+Pose}}
\providecommand{\armDet}{\textsc{Det}}
\providecommand{\armProb}{\textsc{Prob}}
\providecommand{\armProbC}{\textsc{Prob+Com}}
\providecommand{\armDetM}{\textsc{Det+Margin}}

\section{Evaluation}
\label{sec:evaluation}

The stability field of Sec.~\ref{sec:impl_stability} has two consumers, the
robot planner and the human interface.
The simulation study asks whether a twin that samples its physical parameters
selects safer placements than one that uses point estimates (Q1); the user
study asks whether people who see the field place objects more successfully
(Q2).

\subsection{Simulation study}
\label{sec:sim}
\label{sec:ablation}   

\subsubsection{Scenarios and failure criterion}
\begin{figure}[t]
  \centering
  \includegraphics[width=\linewidth]{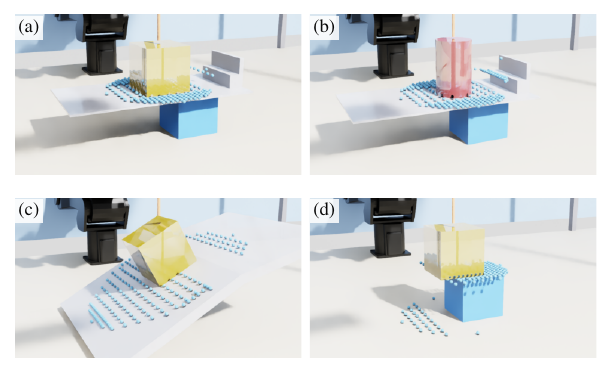}
  \caption{Four Isaac twin scenarios at representative sweep settings.
  Blue dots mark the fixed release candidates shared by all conditions; all
  cubes are \SI{50}{\milli\metre}.
  \textbf{(a,b)} Seesaws: plate COM $10/6\,\si{\milli\metre}$
  inside the supporting cube's pivot edge; cube/cylinder riders land
  $r=10/11/12$ or $4/5/6\,\si{\milli\metre}$ beyond it,
  respectively (largest shown). Tipping depends on $r$ relative
  to the balance point.
  \textbf{(c)} Kinked ramp: $15^\circ$ upper face,
  $\theta_2\in\{26,29,32\}^\circ$ lower face ($29^\circ$ shown),
  and release \SI{30}{\milli\metre} downslope. Sliding depends on
  $\mu_{\text{pair}}$ versus $\tan\theta_2=0.488/0.554/0.625$
  (nominal $\mu=0.60$).
  \textbf{(d)} Cube-on-cube: COM offset
  $20/25\,\si{\milli\metre}$ ($25$ shown, at the edge);
  geometry determines stability.}
  \label{fig:scenes}
\end{figure}

Eleven settings across four scenarios each isolate one failure mode
(Fig.~\ref{fig:scenes}): a seesaw whose rider tips the plate, a kinked ramp
on which the object slides, and a cube-on-cube stack that fails on geometry
alone.
A placement fails if any body translates more than \SI{30}{\milli\metre} or
rotates more than $20^\circ$ within \SI{3}{\second} of release and does not do
so in the matched baseline rollout of Sec.~III-B.
We use this label rather than the kinetic-energy consequence of Sec.~III-B
because three engines must score it identically and static-friction creep is
engine-specific; Eq.~\eqref{eq:mc} becomes the fraction of rollouts that
collapse.

\subsubsection{Compared conditions}
All conditions score the same fixed set of candidate poses around the request
and select
\begin{equation}
  \hat{c} = \operatorname*{arg\,min}_{c \in \mathcal{C}} \; \mathrm{pred}(c)
            + \lambda\,\frac{d(c)}{R},
  \label{eq:rule}
\end{equation}
where $\mathrm{pred}(c)$ is the predicted collapse rate, $d(c)$ the distance to
the request and $R$ the search radius.
With $\lambda{=}0.15$ distance never outweighs one level of predicted risk,
so each condition takes the nearest candidate it predicts safest.
\armReq{} keeps the requested pose.
\armGeo{} is a frictionless support-polygon test at nominal mass, and
\armGeoK{} convolves it with a \SI{4}{\milli\metre}, $2^\circ$ pose kernel,
the landing noise the evaluation injects.
\armProb{} is the proposed twin: it draws $K{=}4$ realizations of mass,
friction, restitution and damping per body and simulates each.
The belief is a clipped normal per parameter: mass at its registry value
$\pm\SI{12}{\percent}$, friction $0.60\pm\SI{16}{\percent}$, restitution
$0.05\pm\SI{12}{\percent}$, damping $0.15/0.25\pm\SI{12}{\percent}$; contact
friction is the mean of the two bodies' draws.
\armDet{} is the same twin with the belief collapsed to its mean, so its
rollouts coincide and Eq.~\eqref{eq:mc} has zero variance; only sampling
differs.
\armProb{} predicts in $\{0,0.25,\dots,1\}$ and \armDet{} in $\{0,1\}$; we
test below whether this resolution alone explains the gap.
The drop test adds two conditions: \armProbC{}, which also samples a
centre-of-mass offset (\SI{2.5}{\milli\metre} SD per axis), and \armDetM{},
\armDet{}'s map with every candidate within \SI{10}{\milli\metre} of a
predicted collapse also marked unsafe.

\subsubsection{Protocol and statistics}
Trials run in the Isaac~Sim twin of Sec.~\ref{sec:impl_twin}, with the
proposing simulator (Warp XPBD~\cite{macklin2016xpbd,warp2022}) and the scoring
simulator (PhysX~\cite{isaacsim}) kept separate and a simulated perception
bias on the planned pose.
Each setting is run 60 times per condition; one sliding trial is excluded by a
rule fixed in advance, leaving 360 tipping and 179 sliding pairs.
A MuJoCo drop test without the arm~\cite{todorov2012mujoco} (100 trials over
ten seeds) hosts all sensitivity analyses reported below.
The primary contrast and its analysis were fixed in a hashed protocol before
any trial and computed blind to condition; \armGeoK{} and the sensitivity
analyses were added later, their analysis frozen before their data were seen.
We report paired differences per failure mode, never pooled, with
\SI{95}{\percent} nested cluster-bootstrap intervals over ten belief
seeds~\cite{field2007clusterbootstrap} and Holm correction~\cite{holm1979}.

\subsubsection{Results}

\begin{table}[t]
\caption{Failure rate (\%), robot in the loop, $K{=}4$ rollouts.
Tipping: 360 paired trials; sliding: 179. Brackets: \SI{95}{\percent}
nested-bootstrap intervals; a dash marks a condition identical to \armReq{} on
every trial.}
\label{tab:ablation}
\centering
\footnotesize
\setlength{\tabcolsep}{3pt}
\resizebox{\columnwidth}{!}{%
\begin{tabular}{@{}lll@{}}
\toprule
condition (belief)                & tipping & sliding \\
\midrule
requested pose, uncorrected       & $38.6$ $[33.3,43.9]$ & $32.9$ $[27.2,38.4]$ \\
\armReq{}\; reactive, no field    & $31.9$ $[24.7,39.4]$ & $32.9$ \;--- \\
\armGeo{}\; shape                 & $25.0$ $[18.3,31.9]$ & $32.9$ \;--- \\
\armGeoK{}\; shape + pose         & $0.3$                 & $32.9$ \;--- \\
\armDet{}\; physics, point        & $25.8$ $[19.7,32.5]$ & $16.2$ $[10.0,23.0]$ \\
\textbf{\armProb{}\; physics, sampled (ours)} & $18.3$ $[12.2,25.0]$ & $11.7$ $[6.1,17.8]$ \\
\quad executed open loop          & $32.5$                & $1.1$ \\
\midrule
\armProb{}$-$\armDet{}            & $-7.5$ $[-11.1,-4.2]$ & $-4.5$ $[-9.4,-1.1]$ \\
\quad Holm-adjusted $p$           & $0.0004$              & $0.015$ \\
\quad trials chosen differently   & $30$ vs.\ $3$         & $8$ vs.\ $0$ \\
\quad re-scored (PyBullet / soft) & $-7.5$ / $-6.4$       & $-4.5$ / $-13.6$ \\
\quad selected pose, open loop    & $-6.7$                & $-4.5$ \\
\quad single engine (Warp)        & $-10.0$               & $-3.8$ \\
\bottomrule
\end{tabular}}
\end{table}

\paragraph{Sampling versus a point estimate (Q1)}
\armProb{} (Table~\ref{tab:ablation}) fails less often than \armDet{} on both
failure modes, by \SI{7.5}{pp} on tipping and \SI{4.5}{pp} on sliding; the sign holds when
outcomes are re-scored in PyBullet~\cite{coumans2021pybullet} and soft-contact
MuJoCo, when the selected pose is executed open loop, and when proposing and
scoring run in a single engine so no threshold disagreement is possible.
Ten releases failed to execute, all in \armProb{} conditions at relocated
candidates, and count as failures.
The gain is carried by the graded prediction: in the drop test without
landing noise, \armProb{}'s map thresholded to $\{0,1\}$ is
indistinguishable from \armDet{} ($-0.9$, $+0.4$\,pp), and \armProb{} beats
its own thresholded map by \SI{9.7}{pp} $[-14.4,-5.1]$ on tipping and
\SI{4.8}{pp} $[-9.8,-1.3]$ on sliding.
More samples keep paying on tipping (36.2, 25.7, \SI{15.3}{\percent} at
$K{=}4$, 8, 16) and saturate on sliding by $K{=}4$ (2.2, 1.6,
\SI{1.0}{\percent}); $K{=}4$ is the live service's budget.
The advantage grows when the belief is wrong in the dangerous direction:
true friction at $0.42$ against an assumed $0.60$ raises the sliding gain to
\SI{18.9}{pp}, and true friction at $0.78$ lowers it to \SI{2.6}{pp}
$[-4.5,-1.0]$ without reversing it.

\paragraph{Geometric versus physical uncertainty}
On sliding every geometric condition matches \armReq{}, because a support
polygon carries no information about friction.
On tipping the order reverses: \armGeoK{} fails on 1 of 360 trials against
\armProb{}'s 66, a difference of \SI{18.1}{pp} $[12.2,24.4]$.
Both estimate the rider's offset from the balance point, on which a
\SI{12}{\percent} mass prior puts \SI{2.0}{\milli\metre} of standard
deviation and pose error 4--\SI{6}{\milli\metre}.
The kernel acts as a safety margin rather than a noise model: a
\SI{2}{\milli\metre} kernel removes the tipping failures even with no noise
injected, and \armDetM{} does the same (0.0 against \armProb{}'s
\SI{36.2}{\percent} without landing noise; 4.2 against 38.5 under
\SI{4}{\milli\metre} noise)
while leaving sliding unchanged, where only sampled friction helps.
A kernel below the true noise loses 9--\SI{11}{pp}; a larger one costs nothing
measurable, consistent
with~\cite{weisz2012posererror,drigalski2022uncertaintyaware}.
Sampling the centre of mass matters most on tipping: in a drop test whose true
COM is also uncertain, \armProbC{} fails \SI{16.8}{\percent} against 40.5
for \armProb{} and 50.1 for \armDet{} (\SI{3.6}{\percent} at $K{=}8$), leaves sliding
untouched, and relocates a median \SI{10}{\milli\metre}; \SI{20}{pp} of the
gain persists when the world's COM is fixed, so most of it is again a margin
bought with extra variance.
Across landing noise of 0, 4 and \SI{8}{\milli\metre}, sampled physics beats
the geometric map on tipping only at zero noise (\SI{10.6}{pp}) and is
overtaken by 34--\SI{42}{pp} from \SI{4}{\milli\metre} on, whereas on
sliding it is the only term that moves ($-3.7$ to \SI{-6.3}{pp}).

\paragraph{Combining the two}
Adding the pose variable to Eq.~\eqref{eq:expectation} costs no extra
rollouts.
Under \SI{4}{\milli\metre} landing noise, convolving the frozen \armProb{} map
with the kernel cuts tipping failures from 38.5 to \SI{0.5}{\percent} and
leaves sliding unchanged, where sampling still contributes \SI{6.1}{pp} over
the equally convolved \armDet{}.
At $0.5/2.9/\SI{0.0}{\percent}$ across the three failure modes it is the only
online-capable condition at or below \SI{3}{\percent} on all of them.

\paragraph{Summary}
The simulation study answers Q1 with a qualification.
A twin that samples its physical parameters selects safer placements than a
point-estimate twin on both failure modes, and the gain is the propagated
uncertainty itself rather than the resolution of its predictions.
The two failure modes, however, call for different uncertainty: tipping is a
margin problem that pose noise, a dilated map or a wider belief all solve,
whereas sliding is a knowledge problem that only sampled friction solves.
Because Eq.~\eqref{eq:expectation} takes both as sampled variables in one
expectation, the field carries each at no extra cost, and the composed
condition is the only one at or below \SI{3}{\percent} on every failure
mode.

\subsection{User study}
\label{sec:study}

\subsubsection{Participants and design}
\begin{figure}[!htbp]
  \centering
  \includegraphics[width=\columnwidth]{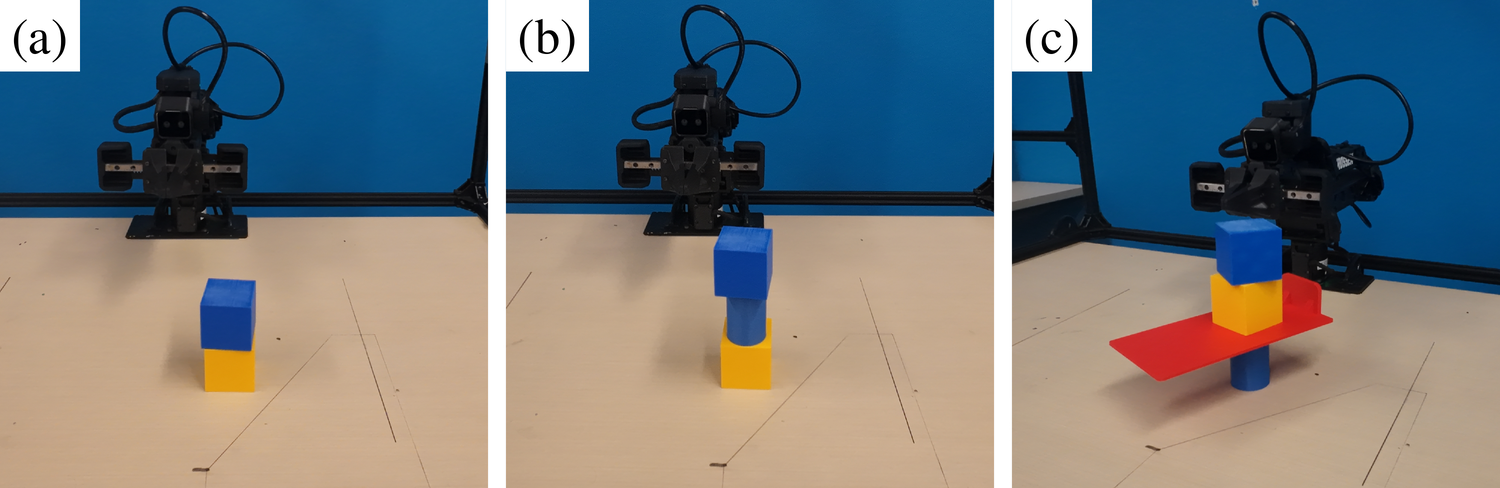}
\caption{User study stacking tasks.
\textbf{(a)} Simple stack (T1).
\textbf{(b)} Complex stack (T2).
\textbf{(c)} Balanced stack (T3).}
\label{fig:user-study-setup}
\end{figure}

Eight participants (P1--P8) took part in a preliminary within-participant
study with two interface conditions: probabilistic guidance (P) shows the
stability heat map and applies the placement guidance of
Sec.~IV-C; direct placement (D) disables both.
Each participant performed the three stacking tasks of
Fig.~\ref{fig:user-study-setup} in both conditions, 48 trials in
total, camera view and
robot execution were identical, and condition order was counterbalanced
across participants.

\subsubsection{Procedure and measures}
After giving consent, participants performed T1 to T3 in sequence with no
practice trials.
A trial succeeded when the complete arrangement was reached; a collapse during
an intermediate placement counted as failure.
A questionnaire and a short interview followed.

\subsubsection{Results (Q2)}
Participants reached the target arrangement in 19 of 24 trials
(\SI{79.2}{\percent}) with guidance and 17 of 24 (\SI{70.8}{\percent}) without.
The only notable difference is on T3, the only task that can tip, where four
participants succeeded with guidance and two without.
Three participants did better with guidance, one worse, four the same.


\begin{figure}[t]
\centering
\includegraphics[width=\columnwidth]{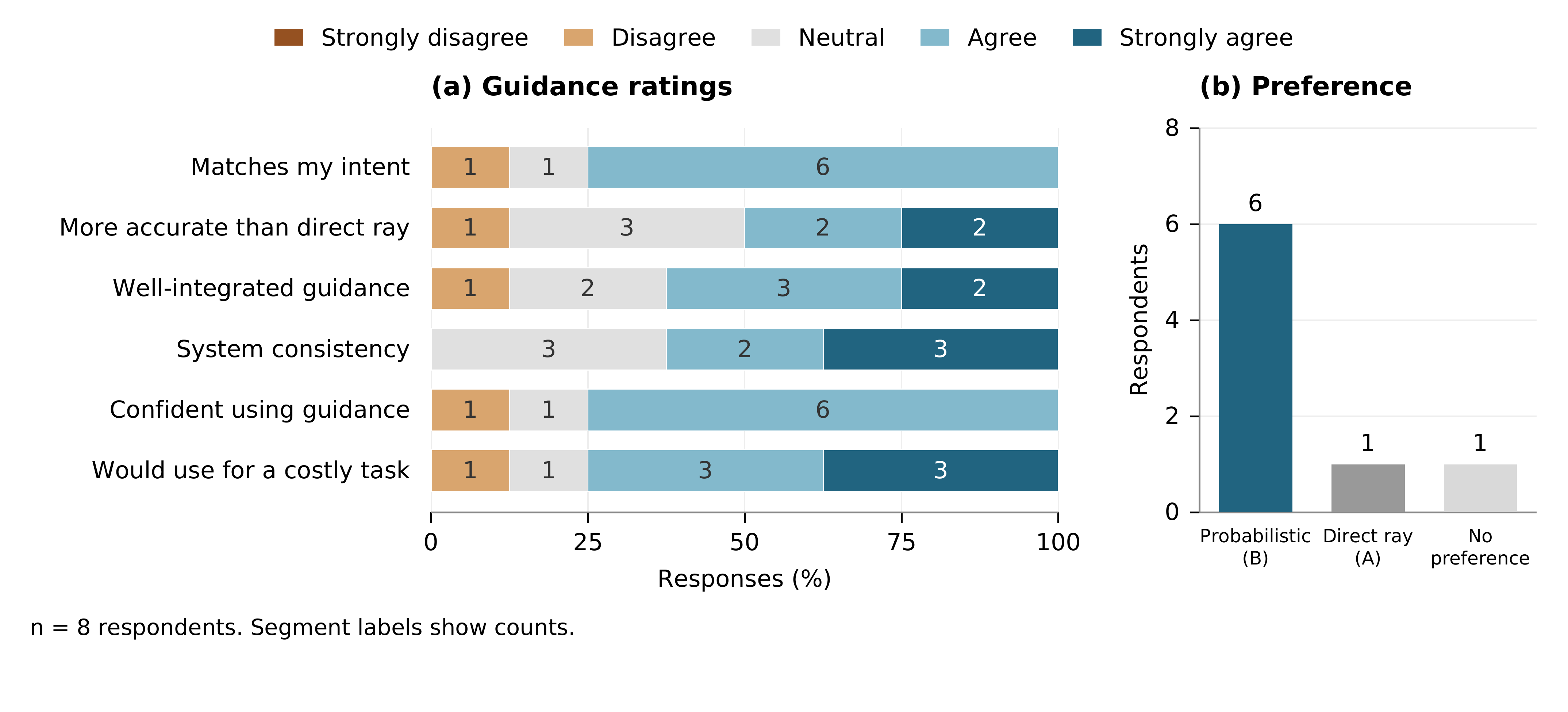}
\caption{Post-task questionnaire results from eight participants.
\textbf{(a)} Guidance ratings.
\textbf{(b)} Interface preference.}
\label{fig:user-study-questionnaire}
\end{figure}

In the questionnaire (Fig.~\ref{fig:user-study-questionnaire}), six
participants preferred guidance, one direct placement, one neither; six agreed
or strongly agreed they would use guidance when mistakes are costly.
In interviews, P1 valued guidance on unfamiliar tasks, P4 found it intuitive,
and P7 used it to check planned placements and preferred it despite fewer
successes in that condition.
With eight participants and a two-trial difference the study is descriptive
only, and condition P combines visualization with automatic snapping, so it
cannot separate what people saw from what the system corrected.
It gives initial evidence for Q2 on the tipping task.

%% file: tex/6-discussion-limitation.tex
\section{Limitations and Future Work}

We present a probabilistic physics framework for Digital Twins that supports
human guidance and robotic action selection. Several limitations remain and
motivate future work.

First, the posterior refinement described in Sec.~\ref{sec:posterior} is
currently theoretical and is not implemented or evaluated in the present
system. Evaluating such updates in the real world is challenging because the
underlying physical parameters are often difficult to measure accurately. In
future work, we plan to evaluate posterior refinement through its predictive
utility, for example by measuring whether the updated model reduces the
discrepancy between simulated and observed real world outcomes.

Second, our evaluation of robotic action selection is currently conducted in
simulation. Although this enables controlled comparisons across uncertainty
conditions, future work should validate the same planning benefits on a physical
robot under real sensing, actuation, and model mismatch.

Finally, we provide a detailed implementation and evaluation only for placement
stability. Impact risk and other task-specific consequence
functionals are currently used to illustrate the generality of the framework.
Future work will investigate their full implementation and study how multiple
consequence fields can be combined to support richer human--robot collaboration.